\documentclass[conference]{IEEEtran}
\IEEEoverridecommandlockouts
\usepackage{cite}
\usepackage{amsmath,amssymb,amsfonts}
\usepackage{graphicx}
\usepackage{textcomp}
\usepackage{xcolor}
\usepackage{times}
\usepackage{booktabs}
\usepackage{multirow}
\usepackage{latexsym}
\usepackage{mathrsfs}
\usepackage{textcomp}
\usepackage{xcolor}
\usepackage{diagbox}
\usepackage{hyperref}
\usepackage{algpseudocode}
\usepackage[ruled]{algorithm2e}
\def\BibTeX{{\rm B\kern-.05em{\sc i\kern-.025em b}\kern-.08em
    T\kern-.1667em\lower.7ex\hbox{E}\kern-.125emX}}
\begin{document}

\title{Enhancing Event Causality Identification with Rationale and Structure-Aware Causal Question Answering}
\author{
    \IEEEauthorblockN{Baiyan Zhang, Qin Chen$^*$, Jie Zhou, Jian Jin, Liang He}
    \IEEEauthorblockA{\textit{School of Computer Science and Technology} \\
        \textit{East China Normal University}, Shanghai, China \\
        byzhang@stu.ecnu.edu.cn \quad 
        \{qchen, jzhou, jjin, lhe\}@cs.ecnu.edu.cn}
}

\maketitle

\begin{abstract}
Document-level Event Causality Identification (DECI) aims to identify causal relations between two events in documents. Recent research tends to use pre-trained language models to generate the event causal relations. Whereas, these methods are prone to the errors of sequential generation due to multiple events in a document. Moreover, the potential structures such as event coreference and related causal chain are neglected. In this paper, we propose a multi-task learning framework to enhance event causality identification with rationale and structure-aware causal question answering. Specifically, the DECI task is transformed into multiple-choice question answering, and the causes and effects of the questioned event are generated with large language models. In addition, we generate the rationales to explain why these events have causal relations. Moreover, we construct an event structure graph, which models the multi-hop potential relations for causal reasoning of the current event. Experiments on two benchmark datasets show the great advantages of our proposed approach compared to the state-of-the-art methods. Moreover, we conduct both quantitative and qualitative analyses, which shed light on why each component of our approach can lead to great improvements.
\end{abstract}


\section{Introduction}
\begin{figure}
\centering
\includegraphics[width=0.5\textwidth]{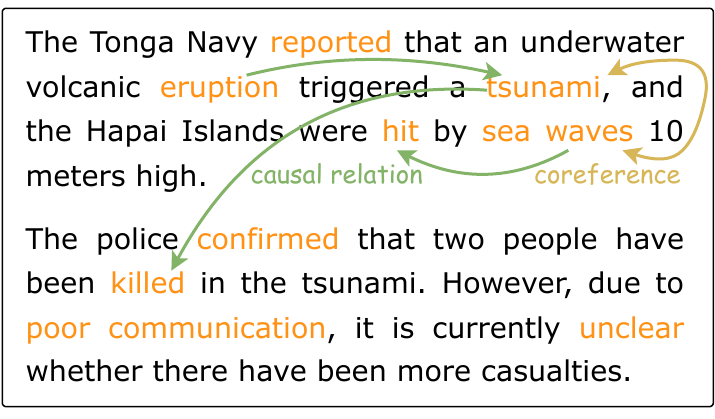}
\caption{A brief example of DECI. The orange words mark all the events. Green lines denote target causal relations and yellow lines denote coreference. This figure does not indicate all the relations in this short text.} \label{fig1}
\end{figure}

Document-level Event Causality Identification (DECI) aims to identify causal relations between two events in documents and it is a crucial and difficult task in information extraction. The research on causal relation extraction helps us to solve many practical problems, such as the automatic processing of massive information, and then support rich downstream applications such as future event forecasting, knowledge graph construction and global crisis monitoring \cite{phu2021graph, li2021causality}. As shown in Fig.~\ref{fig1}, the event \textit{`eruption'} caused \textit{`tsunami'}, so it can be considered that these two events have a \textbf{causal} relation. The event mention \textit{`sea waves'} and event mention \textit{`tsunami'} refer to the same event, so it can be considered as a \textbf{coreference} relation.

\par
Recent advancements in large language models (LLMs) have generated considerable interest in applying these models to various language tasks, such as model editing \cite{meng2022locating, hartvigsen2022aging} model distillation \cite{hsieh2023distilling, jiang2023lion} and parameter-efficient fine-tuning \cite{houlsby2019parameter, hu2021lora}. However, within the task of causal event extraction, recent research tends to use pre-trained language models to generate the event causal relations. Whereas, these methods are prone to errors of sequential generation, limited performance, and high costs due to multiple events in a document. In the era of rapid development of large language models, the potential of these traditional approaches has been severely limited. Even with the addition of components, achieving significant breakthrough performance seems to be a daunting challenge.

GenECI \cite{man2022event} uses a generative model (T5) to infer causal relations between events by capturing key contextual words. However, GenECI uses true or false questions to determine causal relations between two events, which results in high time costs and is only suitable for sentence-level extraction. SEAG \cite{tao2023seag} unifies event detection and event causality identification and proves the effectiveness of linearizing event causal graphs (ECG) through linearization and contrastive learning methods. However, SEAG is not suitable for causal extraction with labeled events, and labeling event words or re-mentioning events separately are difficult to improve its performance in causal extraction with known events effectively. Meanwhile, existing generative methods only generate results through supervised templating and the language model may not comprehend why such extraction is correct.

Unlike traditional methods based on encoder-only pre-trained models and Graph Neural Network variants \cite{zhao2021document, phu2021graph, chen-etal-2023-cheer}, we focus on important research directions in the current Natural Language Processing field and use decoder-only's generative large language model to transform causal relation extraction into a task that can be solved through natural language generation while adapting to limited computational costs. Specifically, we transform causal identification between event pairs into a multiple-choice question. For each event, we refer to it as an "observed event", asking which of the listed options are its cause or effect, and combining reliable rationales provided by smarter models with the linearized event causality diagram we constructed, aiming to extract important contextual key information and enhance important event relation mention information as indirect supervision, So as to enhance its selection accuracy. Our method has an acceptable time and computational cost, and the effect is also quite significant.

The main contributions of our work are as follows:

1. We use the decoder-only generative Large Language Model for event annotation to extract known causal pairs for the first time.

2. We transfer causal relation extraction into multiple-choice questions format, improving prediction performance through integrating rationales and linearized event causal graphs.

3. Experimental results on two public datasets are comparable to the current state-of-the-art discriminative models and significantly surpass other generative models.

\section{Related Work}\label{section2}
\subsection{Sentence-level ECI}
Sentence-level ECI aims to identify causal relations between events in sentences or short unstructured texts. SemSIn\cite{hu2023semantic} employs a GNN-based event aggregator to merge structural information centered around events and utilizes an LSTM-based path aggregator to capture structural information associated between two events. GenECI \cite{man2022event} converts ECI into true or false questions, using a generative model (T5) to extract causal relations by capturing key contextual words. 
\subsection{Document-level ECI}
Unlike sentence-level ECI (SECI), document-level ECI (DECI) faces greater challenges such as long text cross-sentence inferring and repeated mention of story plots. DSGCN\cite{zhao2021document} uses graphic edge prediction to capture information transmission and interaction between causal events. RichGCN \cite{phu2021graph} constructs a multi-level interaction graph to capture the relevant connections between important objects of DECI in the input document. COLA \cite{wang2023cola} objects rich incidental supervision from temporality and balances covariates from multiple timestamps to remove confounding effects. ERGO \cite{chen2022ergo} and CHEER \cite{chen-etal-2023-cheer} constructed a relation graph to model the different relations between events and event pairs, capturing high-order interactions among event pairs. SENDIR \cite{yuan-etal-2023-discriminative} proposes a new discriminative inference with sparse event representation. SEAG \cite{tao2023seag} integrates event detection and DECI. the authors propose a novel structure-aware event causal relation generation and design a new ECG linearization.

\begin{figure*}
\centering
\includegraphics[width=1.0\textwidth]{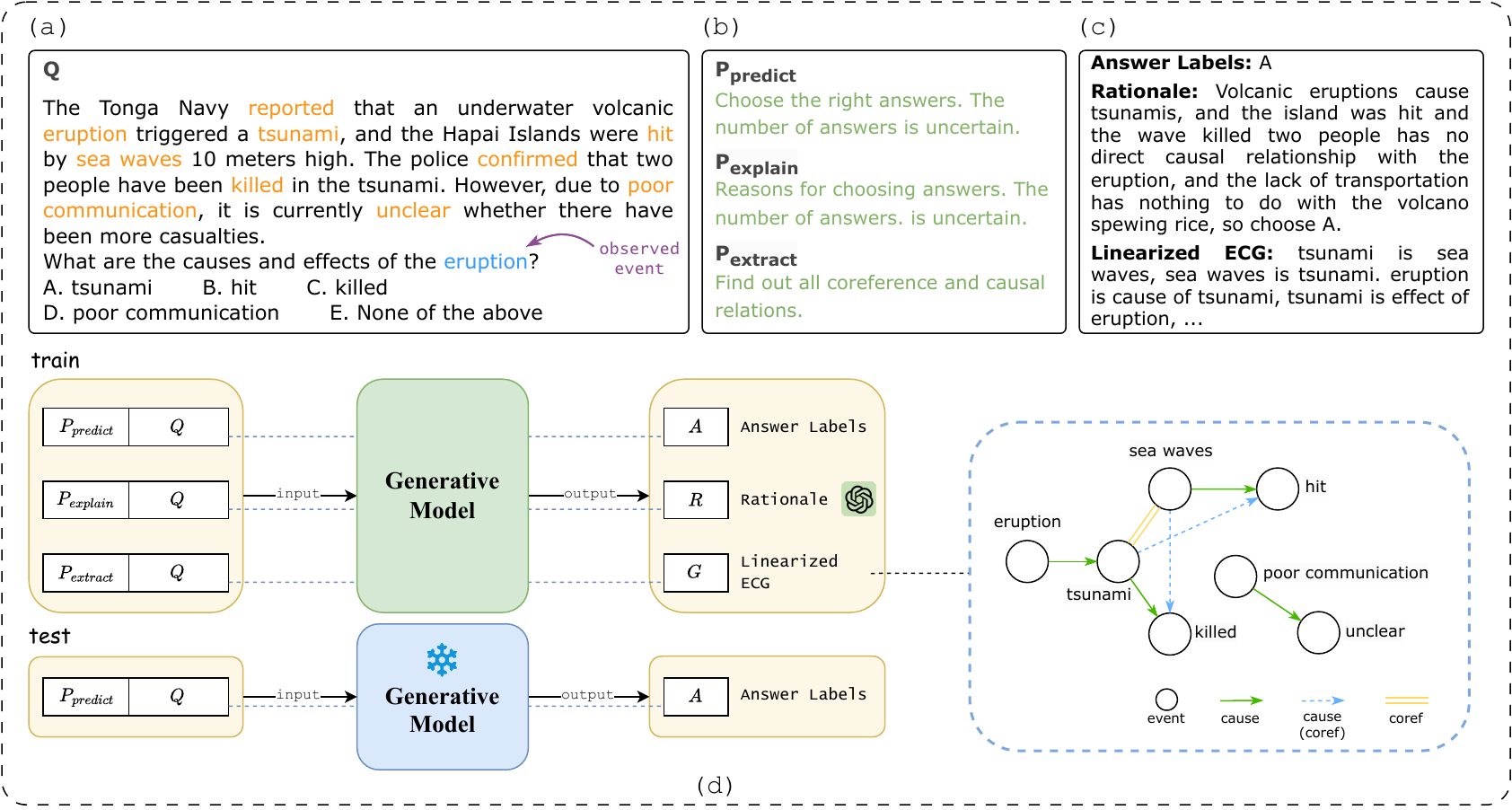}
\caption{An overview of our proposed MCQ Answering integrates Rationale and linearized ECG.} \label{fig2}
\end{figure*}

\subsection{Model Distilling}
The research on large language models has recently become hot, with many pre-trained LLMs of different scales suitable for various tasks proposed, such as LLama, Falcon, Baichuan, etc. Based on these pre-trained LLMs, many research hotspots have emerged, such as model distillation, model editing, parameter efficient fine-tuning, and a series of other works. 
Lora\cite{hu2021lora} freeze the pre-trained model weights and inject trainable low-rank decomposition matrices into every layer of the transformer architecture, significantly reducing the trainable parameter count for downstream tasks. With this approach, it enables us to fine-tune large language models on limited resources.
Human rationales can be used as gold standard labels to make models more explainable by generating similar rationale \cite{narang2020wt5, wiegreffe2020measuring}. Unfortunately, it is expensive to obtain human rationales. Today's LLMs are capable of explaining their predictions by generating high-quality reasoning steps \cite{wei2022chain, kojima2022large}. PINTO \cite{wang2022pinto} releases an LLM to generate rationality at train time and test time, improves the accuracy of model investment, but cannot solve deployment issues. Distilling step by step \cite{hsieh2023distilling} extracts LLM rationales as additional supervision for small models within a multitasking training framework, no additional rationales are required during test time.
\section{Method}
The ECI task refers to inferring all event pairs with causal relationships given a document and all annotated events. We transform it into a multiple-choice question format. Then, the task description is now:
Given the cropped document $\mathbb{D}$ and an observed event $e_i$, To predict the set of events in $\mathbf{E_i}$ that have a causal relationship with the observed event $e_i$ in the set of candidate events in $\mathbb{E}$. The form of the problem is shown in Fig.~\ref{fig2}(a). The main components we propose are shown in Fig.~\ref{fig2}(d).

\subsection{Dataset Processing}
Processing the source document into multiple-choice questions is divided into two parts: text construction and option construction.
\subsubsection{Text Clipping}

Unlike the discriminative models, which directly concatenate event embeddings and put them into a trained classifier for classification, MCQ observes an event and discovers its causal relation with other events, while the development of the story plot often contains causal relations. Therefore, we crop the text, including only the head and tail trimming of observed events and candidate events, to reduce redundant text noise and avoid the influence of irrelevant events, even if this may result in partial semantic loss of text with abnormal narrative order.

Note that using this text clipping method, only the events after the observed event are considered as candidates, and the identification of the relation with the events before the observed event is handled by the previous events. So our question is formulated as \textit{What are the causes and effects of the \{observed event\}?}.

\subsubsection{Option Construction}\label{section3.1.2}
The option is a certain set of candidate events $\mathcal{D}$. For the training set, we add all events with causal relations to the options. If there are coreference events, only the closest one to the observed event is selected, and then at least three unrelated events are selected as interference terms. The advantage of doing so is to reduce the length of the text and learn as much as possible about all relevant causal event pairs. At the same time, the interference term can also be seen as a form of negative samples in contrastive learning, promoting the model to distinguish the relation between different candidate events and the observed event. In summary, The options can be described as (1) Related event mentions, (2) Unrelated interference items, and (3) \textit{"None of the above"}.

For the test set, we sequentially iterate through each event as an observed event, with each set of events mentioned as options. The options can be described as (1) candidate events, (2) \textit{"None of the above"}.

We can infer the similarity between the distribution of the training set and the test set. Let the average number of events in a document be $N$, the number of events with causal relations is $X$, and the number of events without any causal relation is $Y = N - X$. Since the training set traverses all events only once, it is easy to determine that the proportion of samples with correct causal events (i.e. the answer is not \textit{"None of the above"}) in their answers is $P_{has rel}^{train} = \frac{X}{N}$.

For the test set, we can list the following equations 
\begin{equation}\label{formula1}
P_{has rel}^{test} = 1- \frac{Y \cdot M + X \cdot \frac{C_{Y}^{|\mathcal{D}|}}{C_{N}^{|\mathcal{D}|}} \cdot M}{N \cdot M}
\end{equation}
$|\mathcal{D}|$ is the number of options . We assume that causally and non-causally related events are evenly distributed throughout the document and $M$ denotes the average sample size of an observed event in the test set. 
\begin{equation} M \approx \frac{1}{N} \sum_{i=1}^{\frac{N}{|\mathcal{D}|}} |\mathcal{D}|\cdot i 
\end{equation} 
$|\mathcal{D}|$ is also almost equal to the number of events for the same sample size. $M$ does not need to be accurately considered because it will be reduced and the Eq.~\eqref{formula1} will simplify to 
\begin{equation}P_{has rel}^{test} = \frac{X}{N} - \frac{X}{N}\cdot\frac{C_{Y}^{|\mathcal{D}|}}{C_{N}^{|\mathcal{D}|}}
\end{equation} 
$C_n^m$ denotes the number of m elements selected for combination from the given n elements and $\frac{C_{Y}^{|\mathcal{D}|}}{C_{N}^{|\mathcal{D}|}}$ denotes the probability that none of the options are irrelevant. When $Y$ is smaller and $|\mathcal{D}|$ is larger, it gets smaller, much smaller than $\frac{X}{N}$.

The proportion of correct causal events in the training and testing sets of the actual dataset processing results is 53.2\% and 46.9\% respectively, which confirms the rationality of our data process method.

This option construction method enumerates and observes each event for extracting causal extraction, which can reduce the time complexity to $O(d_{train} \cdot n)$ in training and $O(d_{test} \cdot n \cdot \frac{n}{|\mathcal{D}|})$ in testing, and $d_{train}$, $d_{test}$ are the size of the training set and the test set, respectively, far less than the $O (d \cdot n^2)$ of the true or false question. Not to mention that during the testing phase, there's only the single task of generating option ID answers.

\subsection{Multi-task Training}
Formally, we divide multitasking training into three types of tasks, distinguishing different tasks through customized prompts. (1) Explicit extracts relevant event option IDs, which we refer to as Q→A. (2) Extracting rationales for LLM as additional supervision aims to rely on the context comprehension ability of larger LLM to provide reference reasoning steps for the model, denoted as Q→R; (3) The linearization of event causal graphs aims to enhance the model's focus on key events. The more causal relations are mentioned, the higher the token attention weight, denoted as Q→G. The specific method implementation is shown in the Fig.~\ref{fig2}.

\subsubsection{Rationale for MCQ}
During the pre-training phase of the model, multiple-choice questions (MCQs) are commonly utilized as a dataset and task format. This endows the LLM with a certain level of understanding and processing capabilities for MCQ tasks, even if it hasn't been pre-trained on causal MCQ data. Additionally, many pre-trained MCQ datasets include parsing, which provides the reasoning behind selecting an answer. Therefore, we can artificially inject rationales. 

Since the ground truth data of rationales are not given by the original datasets, We use more intelligent models to collect them. Formally, for each MCQ, we use the instruction combined the ground truth label: \textit{The known answer to this question has been determined to be \{answer label, e.g. ABC\}. Please provide a reason for choosing this answer. No more than 50 words.} and put it into a large-scale language model like GPT-3.5, then we get the output as the ground truth rationale of the MCQ.

When the model is trained, As a supervisory tool for predicting answers, it has been proven that the effectiveness of multi-task training differs significantly from that of directly generating answers and rationales (see section \ref{section4.3}). We are also interested in the effect of the rationales generated by different large language models on performance, and relevant experiments can be found in Section \ref{section4.6}.

\subsubsection{ECG Linearization}

In a series of previous works mentioned in Section \ref{section2}, it has been demonstrated that graph neural networks are effective in predicting causal relations, as they aggregate the features of adjacent nodes (where a node is an event), which can enhance the judgment of causal transitivity and event coreference resolution in the model to some extent. The ECG is shown in the blue dashed box of Fig.~\ref{fig2}(d) as part of the enhancement representation. It is a graph $G = (\mathcal{E}, \mathcal{V}) $, while $\mathcal{E}$ is a subset of $\mathbb{E}$. $e_i \in \mathcal{E}$ is an event with causal relations and an edge $v_{ij} \in \mathcal{V}$ denotes there exists a causal or coreferential relation between $e_i$ and $e_j$. Taking inspiration from SEAG \cite{tao2023seag}, We sort the events based on their positions. To incorporate event coreference and causality in a more natural language manner, we linearized G as follows:
\begin{equation}
\hat{g} = [e_1, r_{coref}, e_2, ..., e_1, r_{cause}, e_3, ... ]
\end{equation}

First, mentioning all identical events, and then all causal relationships. In our experiments, we take \textit{is} as $r_{coref}$ and \textit{is the cause of} as $r_{cause}$. Each relationship will be repeated in reverse later, such as \textit{$e_1$ is $e_2$, $e_2$ is $e_1$, ..., $e_1$ is the cause of $e_3$, $e_3$ is the effect of $e_1$, ...}. It should be noted that linearized ECG is only related to the split text and not to the observed events and options.

\subsubsection{Training}
The generative model $f$ learns three tasks simultaneously at once and is trained to minimize the following label prediction loss:

\begin{equation}
\mathcal{L} = \sum_{t \in \mathcal{T}} \alpha_t \mathcal{L}_t
\end{equation}

\begin{equation}
\mathcal{L}_{\text {label }}=\frac{1}{N} \sum_{i=1}^N \ell\left(f\left(p_{\text {predict }}, q\right), \hat{a}\right)
\end{equation}

\begin{equation}
\mathcal{L}_{\text {rationale }}=\frac{1}{N} \sum_{i=1}^N \ell\left(f\left(p_{\text {explain }}, q\right), \hat{r}\right)
\end{equation}

\begin{equation}
\mathcal{L}_{\text {ecg }}=\frac{1}{N} \sum_{i=1}^N \ell\left(f\left(p_{\text {extract }}, q\right), \hat{g}\right)
\end{equation}
Where $\alpha$ represents the loss weight between tasks, $\ell$ is the cross-entropy loss between the predicted and target tokens. One task is trained with a customized prompt and question $(p_t, q)$ as input. $t$ is the specific task, $\mathcal{T}$ is the set of tasks: Q→A, Q→R, Q→G, Correspond to the loss $\mathcal{L}_{\text {label }}$, $\mathcal{L}_{\text {rationale }}$, $\mathcal{L}_{\text {ecg }}$, prompt {$p_{\text {predict }}$, $p_{\text {explain }}$, $p_{\text {extract }}$, and output $\hat{a}$, $\hat{r}$, $\hat{g}$ separately. The examples of prompt and output are shown in Fig.~\ref{fig2}(b)(c).

\section{Experiments}

\subsection{Setup}
\subsubsection{Datasets}
We conducted experiments on two widely used datasets, \textbf{EventStoryLine} \cite{caselli2017event} contains 22 topics, 258 documents, 4,316 sentences, 5,334 event mentions, a total of 54,326 event pairs (including 7,805 intra-sentence event pairs and 4,6521 inter-sentence event pairs), and 5,655 event pairs with causal relations (including 1,770 intra-sentence event pairs and 3,885 inter-sentence event pairs). \textbf{Causal-TimeBank} \cite{mirza2014extracting} contains 184 documents and 6,813 events. Among them, 318 events have a causal relation with the annotations. And the relation span is within three sentences, so it can be considered equivalent to intra-sentence relations.

\begin{table*}
\centering
\caption{
Models’ Document level causal extraction performance on EventStoryLine and Causal-TimeBank, the best results are in \textbf{bold} and the second-best results are \underline{underlined}.}
\resizebox{0.8\textwidth}{!}{
\begin{tabular}{l|ccccc|ccc}
\hline
\multirow{2}{*}{\textbf{Model}}  & \multicolumn{5}{c}{\textbf{EventStoryLine}} & \multicolumn{3}{c}{\textbf{Causal-TimeBank}}\\
\cline{2-9} 
& \textbf{P} & \textbf{R} & \textbf{F1} & Intra F1 & Inter F1 & \textbf{P} & \textbf{R} & \textbf{F1}\\
\hline
LR+ & 27.9 & 47.2 & 35.1 & 40.7 & 33.1 & - & - & -\\
LIP & 36.2 & 49.5 & 41.9 & 44.6 & 40.6 & - & - & - \\
BERT & 41.3 & 38.3 & 39.7 & 52.1 & 32.6 & 47.6 & 55.1 & 51.1\\
RichGCN & 42.6 & 51.3 & 46.6 & 55.2 & 42.2 & 39.7 & 56.5 & 46.7\\
ERGO & 43.2 & 48.8 & 48.1 & 59.0 & 45.8 & \underline{62.1} & 61.3 & \underline{61.7}\\
CHEER & \textbf{49.7} & 53.3 & 51.4 & 62.6 & \textbf{48.4} & 56.4 & \textbf{69.5} & \textbf{62.3}\\
SENDIR & 37.8 & \textbf{82.8} & \underline{51.9} & \textbf{66.2} & \underline{48.3} & \textbf{65.2} & 57.7 & 61.2\\
\hline
GPT-3.5 (0-shot) & 13.9 & 54.7 & 22.2 & 35.5 & 16.4 & 27.7 & 55.3 & 36.9\\
GenECI & - & - & - & 58.8 & - & 60.1 & 53.3 & 56.5\\
Ours & \underline{48.7} & \underline{56.0} & \textbf{52.1} & \underline{65.3} & 47.2 & 58.8 & \underline{63.8} & 61.2\\
\hline
\end{tabular}}
\label{tab1}
\end{table*}

EventStoryLine provides a coreference chain of event mentions, but Causal-TimeBank does not. Although Causal-TimeBank provides sufficient text, its relations are sparse and mainly exist within no more than three sentences, leaning more towards SECI. In order to pay more attention to document-level causal event pair discrimination, we will use EventStoryLine for further research in the follow-up.
\subsubsection{Metrics}
Users often do not pay special attention to whether the event pairs are cross sentences or within sentences, but instead provide a paragraph of text to obtain more accurate causal event pairs. Therefore, we choose Precision, Recall, and F1-score as the main evaluation indicators, and attach intra-sentence F1-score and inter-sentence F1-score.
\subsubsection{Baselines}
(1) LR+ and LIP \cite{gao2019modeling}, feature-based methods, construct document-level structures and utilize various types of resources.
(2) RichGCN \cite{phu2021graph}, constructs a multi-level interaction graph to capture the relevant connections between important objects of DECI in the input document.
(3) BERT, a baseline method that utilizes dynamic windowing and event tagging techniques, follows the previous implementation by \cite{chen2022ergo}.
(4) ERGO and CHEER \cite{chen2022ergo, chen-etal-2023-cheer}, construct a relation graph to model the different relations between events and event pairs, capturing high-order interactions among event pairs. 
(5) SENDIR \cite{yuan-etal-2023-discriminative}, proposes a new discriminative inference with sparse event representation.
(6) GenECI \cite{man2022event}, captures key contextual words and infers causal relations in true/false questions.
(7) GPT-3.5\footnote{\url{https://chat.openai.com}}, an excellent and well-known language model. Due to the high training cost, we will only explore its 0-shot capability here

\subsubsection{Implementation Details}
We select 4 events and "None of the above" as the option set $\mathcal{D}$, so $\mathcal{|D|}$ is 5. There might be more options in the training set.

We used the Lora method to perform Parameter-Efficient Fine-Tuning (PEFT) on Baichuan2-7B-Chat\footnote{\url{https://huggingface.co/baichuan-inc/Baichuan2-7B-Chat}}, an open-source large language model fully accessible for academic research purposes, with a Lora rank of 32, epoch of 10, dropout rate of 0.1, and a learning rate of 5e-5. We randomly shuffle and select 80\% as the training set and 20\% as the prediction set and take the average of three experimental results. We use 2 × Nvidia RTX3090-24G GPU, which indicates that even with limited computing power, it still works.

During the training phase, compared to traditional methods with similar performance, our method trains for 10 epochs (about 6 hours) and traditional methods train for 30 epochs (about 3 hours) with the similar settings. In the testing phase, with the same setup, our inference takes about 15 minutes, while the traditional method takes about 5 minutes. Although our time consumption is relatively high, we can still refer to it as "the time and cost is acceptable".

Thanks to LoRA, our checkpoint volume (approximately 67MB) has reached a smaller size than traditional methods for full fine-tuning of traditional methods (over 400MB, without optimizers).

\subsection{Overall Performance}
Table \ref{tab1} shows the overall performance on EventStoryLine and Causal-TimeBank and uses a horizontal line to separate discriminative (top) and generative (bottom) methods.
Our work in document-level causal relation extraction achieves comparable performance to the state-of-the-art CHEER and SENDIR discriminative models on EventStoryLine and Causal-TimeBank. On Causal-TimeBank, we slightly fall short compared to the SOTA, possibly due to the prevalence of relations at the sentence level. The process of constructing options from a limited number of candidate events causes fluctuations in the number of options on the test set, some of which are fewer than $\mathcal{|D|}$. This difference in distribution with the training set results in a loss of performance.
However, we utilized a more user-friendly generative model, which holds greater potential and a higher ceiling in performance. Moreover, it outperforms other generative methods significantly in performance and was notably more cost-effective.

\begin{table}
\centering
\caption{
Ablation study.}
\begin{tabular}{l|ccccc}
\hline
\textbf{Model} & \textbf{P} & \textbf{R} & \textbf{F1} & Intra F1 & Inter F1\\
\hline
Ours (Q→A, R, G) & 48.74 & 55.96 & 52.10 & 65.30 & 47.04 \\
w/o R (Q→A, G) & 47.37 & 55.35 & 51.05 & 64.31 & 46.03 \\
w/o G (Q→A, R) & 45.44 & 58.36 & 51.10 & 65.85 & 45.66 \\
Q→A & 47.31 & 49.93 & 48.58 & 62.65 & 43.93 \\
Q→A$\oplus$R & 36.50 & 23.33 & 28.47 & 34.43 & 26.34 \\
\hline
\multicolumn{6}{l}{$\oplus$ means concatenation, directly generates both an answer and its rationale.}
\end{tabular}\label{tab2}
\end{table}

\subsection{Ablation Study}\label{section4.3}
In order to further analyze the role of each task, we also conducted an ablation study to demonstrate the effectiveness of our main modules. We show the results of the ablation study in Table \ref{tab2}.

We examined the impacts of tasks Q→R and Q→G.
(1) \textbf{Without rationale (w/o R).} In the absence of rationale, there will be a certain decrease in performance, both within and within sentences, with a decrease of about 1 percentage point. This is because we have not yet removed linearized ECG, and the decrease is not particularly significant. 
(2) \textbf{Without linearized ECG (w/o G).} In the absence of linearization ECG, there will be a certain decrease in performance, but there will be a certain enhancement within the sentence and a more significant decrease between sentences. This may be because the rationales generated by the large model focus more on the effect of nearby events and mention less distant events in the rationales.
(3) \textbf{Neither is included (Q→A).} The decrease of more than 3 percentage points proves our reasoning and the effectiveness of linearizing ECG. 
(4) \textbf{Q→A$\oplus$R} refers to the generation of answers and rationales in one conversation, which is surprisingly low, with a decrease of more than 20 percentage points. This reminds us that when training choice question datasets, it is not appropriate to directly generate answers and parse them as targets at once. It cannot obtain the best results, and multitasking the generation of answers and explanations will result in better results.

\begin{figure}[b]
\centering
\includegraphics[width=0.5\textwidth]{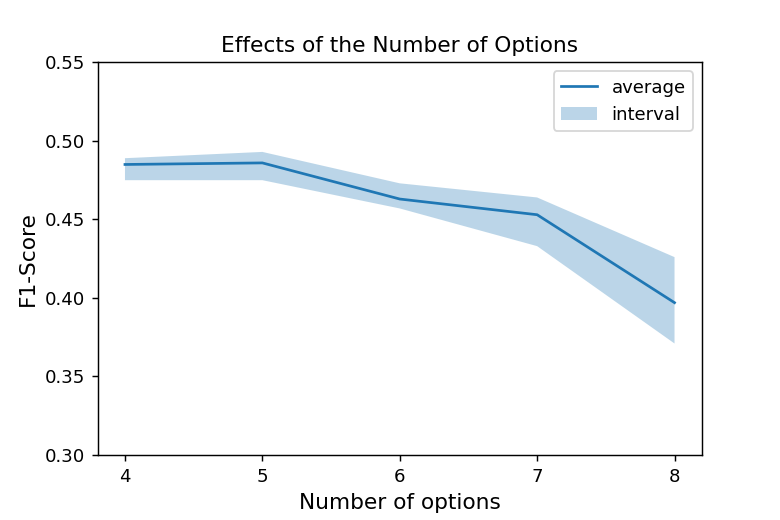}
\caption{Effect of number of options. The blue area represents the deviation interval.} \label{fig3}
\end{figure}

\subsection{Effects of Number of Options}

To better align with the nature of the large-scale model itself and reduce the cost of increasing the number of options, during training, a single task of generating answers (Q→A) was employed to explore the effect of the Number of Options on the outcome. The results are depicted in Fig.~\ref{fig3}. It is demonstrated that utilizing four events along with one related event yields the best performance. In the case of no more than four options, the distribution difference between the test set and the training set is greater, and the number of inference samples is larger, as predicted by section \ref{section3.1.2}. At greater than five, the multiple-choice questions have more options and it is more difficult to select the correct options maybe due to the rarity of pre-trained data.

\begin{table}
\centering
\caption{
Without considering ECI, the effect of different weights. $alpha$ represents the weight of Q→A and the weight of Q→R is $1-alpha$.
}
\small
\resizebox{0.5\textwidth}{!}{
\begin{tabular}{c|ccccc}
\hline
\textbf{alpha} & \textbf{P} & \textbf{R} & \textbf{F1} & Intra F1 & Inter F1\\
\hline
0.7 & 47.51 & 47.86 & 47.70 & 61.54 & 42.40 \\
0.5 & \textbf{48.17} & 54.31 & 51.06 & 63.73 & 46.13 \\
0.4 & 46.68 & 54.93 & 50.97 & 61.15 & 45.53 \\
0.3 & 45.44 & \textbf{58.36} & \textbf{51.10} & 65.85 & 45.66 \\
0.1 & 44.20 & 40.63 & 42.34 & 51.22 & 39.02 \\
\hline
\end{tabular}}\label{tab3}
\end{table}

\begin{figure}
\centering
\includegraphics[width=0.5\textwidth]{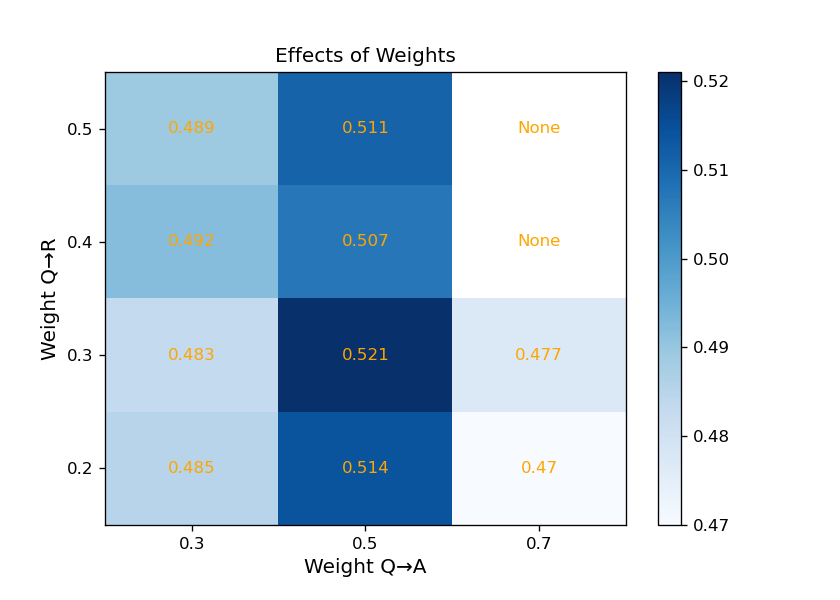}
\caption{Effect of weights of all three tasks. The sum of three task weights is $1.0$. Orange value is F1-score.} \label{fig4}
\end{figure}

\subsection{Effects of Weights}
To streamline task parameter adjustments and focus on the primary task Q→A, we're testing the impact of different Q→A weights on the results. Firstly, we're considering solely Q→A, R. Here, $alpha$ represents the weight of Q→A, while the weight of Q→R is $1-alpha$.

As shown in Table \ref{tab3}. When the Q→A task weight is between 0.3 and 0.5, the F1-score is relatively high, while below 0.3, the model pays too little attention to the key task Q→A, resulting in a loss of performance.

Surprisingly, the performance at $alpha$=0.7 is even lower than that at $alpha$=1 (i.e. single task Q→A), which may be due to the fact that the number of tokens for the rationale is much larger than the number of tokens for the answer, and it occupies a smaller proportion. The model does not have a complete understanding of the rationale's way of thinking, but instead leads to parameter updates that deviate and introduce noise.

And Then, we conducted experiments with limited weight combinations on all three tasks and the results are shown in Fig.~\ref{fig4}. The effect is better when the weight of task Q→A is 0.5.

\subsection{Effects of Generated Rationales}\label{section4.6}
In order to minimize the impact of task weights as much as possible, we did not consider linear ECG tasks and only used a Q→A, R two-task approach, using GPT-3.5-turbo and unfine-tuned Baichuan2 as LLM for generating rationales. The results are shown in Table \ref{tab4}. This implies that as the model's capacity strengthens, it becomes more proficient in generating coherent rationales aligned with the context. When an ample supply of valid justifications is established, the likelihood of further improvement in outcomes increases.
Under a single model, the rationale provided by using the untrained model itself does not improve but impairs performance. This may be because the reasonable rationale generated by the model itself may be more challenging than the correct answer to the causal multiple-choice question, and the unreasonable rationale will degrade the performance of models far more than the reasonable rationale will optimize the performance.
 
\begin{table}
\centering
\caption{
Without considering ECI, the impact of rationales for different LLMs. All LLMs are not fine-tuned.
}
\resizebox{0.5\textwidth}{!}{
\begin{tabular}{l|ccccc}
\hline
\textbf{LLM} & \textbf{P} & \textbf{R} & \textbf{F1} & Intra F1 & Inter F1\\
\hline
GPT-3.5-Turbo & 45.44 & 58.36 & 51.10 & 65.85 & 45.66 \\
Baichuan2-13B & 44.19 & 47.73 & 45.89 & 57.34 & 40.42 \\
Baichuan2-7B & 44.60 & 42.16 & 43.35 & 53.88 & 39.18 \\
\hline
\end{tabular}}\label{tab4}
\end{table}

\subsection{Case Study}
In this section, we conducted a case study to further illustrate an intuitive impression of our components.
In Fig.~\ref{fig5}, We give a text containing several annotated events. The nether table includes two examples of the relation between two events.
For event pair No.1, The rationales can better understand the contextual semantics: \textit{"People throw a projectile, resulting in accidentally injuring a pharmacy customer. "}. Even if the word of the event mention changes and linearized ECG does not work here, the semantic representation with context will still enhance the reasoning ability during the testing phase.
For event pair No.2, Because the distance between the sentence [1] and the sentence [4] is not so close, the rationale often only focuses on the word \textit{`killed'} after the \textit{`anger'}. However, if the ECG is integrated, even if the distance is too long, the long-distance event relation in the training stage will be enhanced, and thus its long-distance reasoning ability will also be enhanced during the testing phase.
However, by integrating the ECG, even if the distance is long, the long-distance event relations learned during the training stage are reinforced. Consequently, this enhances its long-distance reasoning ability during the testing phase.

In fact, another advantage of our approach compared to traditional methods is that when we need to understand why the model get this answer , we can obtain the reasons for its selection by using specific prompts (as shown in Fig.~\ref{fig2}(b)) and questions. This can provide basic rationales or or be used to prompt researchers to solve the shortcomings of the method during reasoning, such as in the rational section of the case study. Therefore, the model's decisions are interpretable.

For the rationale of for the example 1, this means that the model has learned and understood the scenario where a directed projectile may hit a person and cause injury due to a miss. For the second rational, the model provided an incorrect explanation, possibly due to its context spanning multiple sentences. This prompted our method to focus more on handling long text situations, which also corresponds to our proposed linearized ECG method.

\begin{figure}
\centering
\includegraphics[width=0.5\textwidth]{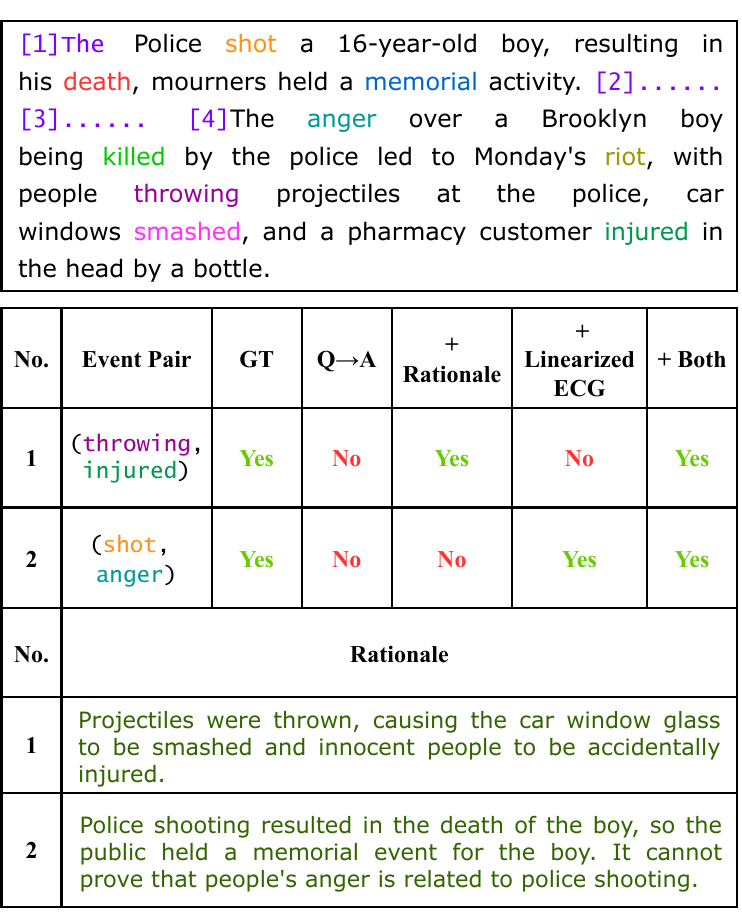}
\caption{A case study. GT denotes ground truth. } \label{fig5}
\end{figure}

\section{Conclusion}
We propose a document-level event causality identification framework based on a generative large language model. This framework transforms the task into a multiple-choice question format, integrating rationales generated by the larger model and linearized event causal graphs, utilizing both multi-task training and single-task inference. We conduct extensive experiments on two commonly used benchmark datasets, achieving results similar to the previous state-of-the-art discriminative model, significantly surpassing other generative models. This work establishes a new baseline for future research in event causality identification by generative models.

\section*{Limitations}
LLM has label ID bias \cite{zheng2023large}, and multiple-choice questions without an ID will to some extent affect the accuracy of answers. Meanwhile, in causal MCQ data, the answer \textit{"E: None of the above"} is numerous, therefore, the label ID bias of MCQ used in this article cannot be ignored.

The construction of option interference terms. By analyzing error cases, it is often predicted that the wrong options are not common options suitable for various scenarios, such as \textit{reported} and \textit{concerned}, but other major events related to the story plot in the text. These major events are not related to the observed events but often have other causal relations. Thus, the construction of options can also be considered as an optimization method.

\bibliographystyle{ieeetr}
\bibliography{custom}
\vspace{12pt}

\end{document}